\documentclass[runningheads]{llncs}

\usepackage[T1]{fontenc}
\usepackage{graphicx}
\usepackage{amsmath}
\usepackage{amssymb}

\usepackage{array}
\usepackage{booktabs}
\usepackage{makecell}

\usepackage{algorithm}
\usepackage{algorithmic}

\usepackage{todonotes}

\usepackage[table]{xcolor}
\usepackage{pgf}

\newcommand{\DeltaR}[1]{%
  \pgfmathsetmacro{\val}{#1}%
  \pgfmathsetmacro{\absval}{abs(\val)}%
  \pgfmathsetmacro{\intensity}{min(85, max(25, round(85*sqrt(\absval))))}%
  \ifdim \val pt > 0pt
    \textcolor{green!\intensity!black}{#1}%
  \else
    \textcolor{red!\intensity!black}{#1}%
  \fi
}

\begin{document}
\title{Street-Legal Physical-World Adversarial Rim for License Plates}
\author{Nikhil Kalidasu\inst{1}\orcidID{0009-0001-2651-8292} \and
Sahana Ganapathy\inst{1}\orcidID{0009-0007-1823-1324}} 
%
%\authorrunning{N. Kalidasu et al.}
\institute{The University of Texas at Austin, Austin TX 78705, USA}
\maketitle
\begin{abstract}
Automatic license plate reader (ALPR) systems are widely deployed to identify and track vehicles. While prior work has demonstrated vulnerabilities in ALPR systems, far less attention has been paid to their legality and physical-world practicality. We investigate whether low-resourced threat actors can engineer a successful adversarial attack against a modern open-source ALPR system. We introduce the Street-legal Physical Adversarial Rim (SPAR), a physically realizable white-box attack against the popular ALPR system \texttt{fast-alpr}. SPAR requires no access to ALPR infrastructure during attack deployment and does not alter or obscure the attacker's license plate. Based on prior legislation and case law, we argue that SPAR is street-legal in the state of Texas. Under optimal conditions, SPAR reduces ALPR accuracy by 60\% and achieves an 18\% targeted impersonation rate. SPAR can be produced for under \$100, and it was implemented entirely by commercial agentic coding assistants. These results highlight practical vulnerabilities in modern ALPR systems under realistic physical-world conditions and suggest new directions for both attack and defense.
\end{abstract}

\section{Introduction}

Automatic license plate reader (ALPR) technology is extensively deployed by governments, communities, and businesses to identify and track vehicles \cite{kocerArtificialNeuralNetworks2011,finkleaLawEnforcementTechnology2024,steinAutomatedLicensePlate2023}. Neural networks have long been used for license plate recognition \cite{liReadingCarLicense2016,liuConvolutionalNeuralNetworksbased2018,masoodLicensePlateDetection2017}. Adversarial attacks against vision models have a similarly long history \cite{szegedyIntriguingPropertiesNeural2014,goodfellowExplainingHarnessingAdversarial2015,kurakinAdversarialExamplesPhysical2018}, including attacks on surveillance systems \cite{nguyenPhysicalAdversarialAttacks2023}.

To investigate whether adversarial attacks pose a realistic threat to modern ALPR systems, we conducted a demonstration attack against a single modern open-source ALPR system. We sought to both prevent the correct reading of the attacker's license plate (a non-targeted "disruption" attack) and induce the defender's ALPR system to output an attacker-defined target (a targeted "impersonation" attack). We imposed the following constraints: \textbf{Realistic Access:} The attacker has no access to the defender's ALPR software or hardware during attack deployment. \textbf{Street-legality:} The attack must be street-legal in the state of Texas. \textbf{Simplicity:} The attack must be plausible for a time- and resource-constrained attacker with limited technical knowledge. \textbf{Practicality:} The attack must be practical in the physical world.

We modeled a scenario where the threat actor has acquired model weights and source code of the defender's ALPR system. This white-box assumption follows standard security practice by assuming that attackers may have detailed knowledge of a system’s design \cite{kerckhoffsCryptographieMilitaire1883,fergusonContextCryptography2010}. We designed a highly specific attack for a single vehicle, license plate, and ALPR system. We limited our total costs to \$100 (excluding evaluation) and restricted the size of our training dataset to what could plausibly be collected and labeled by a individual attacker. All code was generated with commercial agentic coding tools to investigate whether large language models (LLMs) can indeed uplift the capabilities of low-resourced attackers \cite{anthropic2025threat}. We prioritize the street-legality constraint because if an attack does not clearly violate existing laws, it may be difficult for law enforcement to mitigate the threat. A practical street-legal adversarial attack on ALPR systems, in the absence of new legislation, could necessitate the adoption of adversarial robustness measures into deployed license plate readers as an imperfect technological solution. We narrow our scope to the single jurisdiction of Texas to make this problem tractable, as regulations vary widely across jurisdictions. To evaluate physical-world practicality, we test our attack under varying distance, perspective, and lighting conditions, which have all been identified in prior literature as challenges to adversarial attacks on vision models \cite{shackBreakingIllusionRealworld2024} and on ALPR systems specifically \cite{zhaRoLMAPracticalAdversarial2019}. We use statistical methods to quantify our approach's dependence on camera distance and viewing angle.

\subsection{Background}

Adversarial attacks on computer vision models can be broadly classified into perturbation attacks \cite{szegedyIntriguingPropertiesNeural2014}, patch-based attacks \cite{brownAdversarialPatch2018}, and spot attacks \cite{qianSpotEvasionAttacks2020}. Goodfellow et al. \cite{goodfellowExplainingHarnessingAdversarial2015} showed that perturbation attacks exploit the approximate linearity of neural networks in high-dimensional input spaces. By adding a bounded input-dependent perturbation, an attacker can induce a disproportionately large change in a model’s internal activations and output relative to the perturbation magnitude. These methods are effective against modern vision models and have been used against ALPR systems in the past \cite{guAdversarialAttacksLicense2020,kwonAdvPlateAttackAdversarially2021,jiamsuchonFeasibilityAttackingThai2023}.

A major limitation of this strategy is that the attacker must modify an image after capture but before model inference, requiring a level of system access that is often impractical. The adversarial patch method instead overlays a small learnable patch onto input images at random locations and under random transformations \cite{brownAdversarialPatch2018}. A common approach is to freeze the victim model’s parameters and optimize the patch under an expectation-over-transformation (EoT) objective \cite{brownAdversarialPatch2018} that enforces robustness to transformations in viewpoint and camera equipment. This has enabled physical-world attacks against image classifiers, object detectors, and other vision systems  \cite{eykholtRobustPhysicalWorldAttacks2018,liuDPatchAdversarialPatch2019,liuBiasBasedUniversalAdversarial2020,xuAdversarialTShirtEvading2020,taoHardlabelBlackboxUniversal} because a printed patch sticker maintains this robustness. However, adhering a patch on top of a license plate may constitute an "attached sticker" interfering with readability, and may be illegal in the state of Texas \cite{texTranspCode504945}. An alternative approach is to define the patch as a rim surrounding the license plate rather than a sticker applied on top of it; this has shown some promise in prior literature \cite{zhangLPLAAdversarialAttack2024} and is the approach we selected for SPAR.

The spot attack occupies a middle ground between perturbations and patches. Spot attacks add small, perceptible patterns to an image resembling naturalistic noise like mud splatters. They have been successfully demonstrated against ALPR systems, including illumination-based and sticker-based designs \cite{zhaRoLMAPracticalAdversarial2019,qianSpotEvasionAttacks2020}. But in practice, the “stealthiness” of spot attacks remains subjective \cite{eykholtRobustPhysicalWorldAttacks2018}, and strategically obscuring parts of a license plate to inhibit its readability has similar legality concerns to vanilla adversarial patches \cite{texTranspCode504945}.

\subsection{Related Work}

\begin{table}[t]
\centering
\caption{Comparison with prior adversarial attacks on ALPR systems. We strictly define "system-level access" as requiring digital alterations after image capture but before model submission, with no potential for extension to the physical world.}
\label{tab:alpr_comparison}
\resizebox{\textwidth}{!}{
\begin{tabular}{lccccccc}
\toprule
\textbf{Approach} 
& \textbf{Attack Type} 
& \textbf{Stage} 
& \textbf{Targeted} 
& \textbf{\makecell{Physical\\World}}
& \textbf{\makecell{Legal\\ (Texas)}}
& \textbf{\makecell{Perspective-\\Robust}}
& \textbf{Access Model} \\
\midrule
Zha et al. \cite{zhaRoLMAPracticalAdversarial2019} 
& Spot
& OCR 
& \checkmark 
& \checkmark 
& \texttimes 
& \checkmark 
& White-box \\

Gu et al. \cite{guAdversarialAttacksLicense2020}
& Perturbation 
& OCR 
& \texttimes 
& \texttimes 
& \texttimes 
& \texttimes 
& System-level \\

Qian et al. \cite{qianSpotEvasionAttacks2020}
& Spot
& OCR 
& \checkmark& \texttimes 
& \texttimes 
& \texttimes 
& Black-box \\

Kwon \& Baek \cite{kwonAdvPlateAttackAdversarially2021}
& Perturbation 
& OCR 
& \checkmark 
& \texttimes 
& \texttimes 
& \texttimes 
& System-level \\

Jiamsuchon et al. \cite{jiamsuchonFeasibilityAttackingThai2023}
& Perturbation 
& OCR 
& \texttimes 
& \texttimes 
& \texttimes 
& \texttimes 
& System-level \\

Zhang et al. \cite{zhangLPLAAdversarialAttack2024}
& Rim 
& Detection 
& \texttimes 
& \checkmark 
& \checkmark 
& \texttimes 
& White-box \\

\midrule
\textbf{SPAR} 
& \textbf{Rim} 
& \textbf{Both} 
& \textbf{\checkmark} 
& \textbf{\checkmark} 
& \textbf{\checkmark} 
& \textbf{\checkmark} 
& \textbf{White-box} \\
\bottomrule
\end{tabular}
}
\end{table}

Perturbation-, patch-, and spot-based attacks have all been previously executed against ALPR systems, in both white- and black-box settings, against different parts of the ALPR pipeline (\textit{detection} of the license plate in an image versus \textit{reading} of the plate number via optical character recognition, OCR). Prior work explores both targeted attacks which cause the ALPR system to misread a license plate as a specific "impersonation target," and non-targeted attacks which cause the ALPR system to produce non-specific incorrect outputs.

Zha et al. \cite{zhaRoLMAPracticalAdversarial2019} conducted a physical-world spot attack targeting the OCR stage, and identified two key challenges: legal requirements to not modify the license plate area and physical-world robustness to variations in lighting, distance, and camera angle. They therefore applied the spots by illuminating the license plate with an attached electronic device. The complexity of this solution violates our simplicity constraint, and an "illuminated device" interfering with plate readability is explicitly noted as illegal in Texas \cite{texTranspCode504945}. Subsequent work largely focused on digital-world attacks against ALPR systems, primarily targeting the OCR stage under assumptions of post-capture access to images \cite{guAdversarialAttacksLicense2020,kwonAdvPlateAttackAdversarially2021,jiamsuchonFeasibilityAttackingThai2023,qianSpotEvasionAttacks2020}. These approaches assume the ability to intercept and modify images after capture, violating our realistic access constraint, and do not attempt physical-world practicality. Zhang et al. \cite{zhangLPLAAdversarialAttack2024} conducted a physical-world patch attack on the detection stage with a rim surrounding the license plate. However, the authors did not take the step of Zha et al. of accounting for lighting, distance, and camera angle, which are known to cause issues for patch attacks specifically \cite{shackBreakingIllusionRealworld2024}. Additionally, their physical-world evaluation was limited to a single ideal-case test image.

There thus remains a literature gap in adversarial attacks on ALPR systems combining realistic access models, street-legality, simplicity for low-resourced attackers, and physical-world practicality. Additionally, all previous attacks focus exclusively on either the license plate detection portion or the OCR portion of the ALPR pipeline; none have attempted to simultaneously attack both. We therefore present the Street-legal Physical Adversarial Rim (SPAR), a printable adversarial license plate rim that successfully disrupts ALPR in both the digital and physical world while satisfying all our constraints.

\section{Methods}

SPAR is an adversarial rim around the license plate (adopted from Zhang et al. \cite{zhangLPLAAdversarialAttack2024}) which targets both the detection and OCR phases. The detection attack minimizes confidence and expands the bounding box to fully encompass the rim; the OCR attack forces \texttt{fast-alpr} to produce incorrect or attacker-defined output. We ensure SPAR's street-legality by mounting it as a printed poster between the license plate and the vehicle, strictly behind all visible areas of the license plate. Under Texas Transportation Code \S\ 504.945 \cite{texTranspCode504945}, a physical border fully outside the plate boundary does not constitute a "coating, covering, or protective substance" under (a)(7), nor an "attached... sticker, decal, emblem, or other insignia" under (a)(6). United States v. Granado held that a license plate frame is not an "insignia" within (a)(6)'s meaning \cite{usVGranado2002}. United States v. Trevino \cite{usVContrerasTrevino2006} and State of Texas v. Johnson \cite{stateVJohnson2007} held that a frame can be illegal if it covers any part of the legible text. SPAR does not cover any text and would not fall under this exception. So we contend that SPAR is street-legal in Texas.

We selected the \texttt{fast-alpr} package \cite{andresAnkandrewFastalpr2025} as our target because of its modern development and high accuracy. A YOLO-v9 object detection model from \texttt{open-image-models} \cite{andresOpenImageModels2025}, specifically, the \texttt{yolo-v9-t-384-license-plate-\\end2end} detection model \cite{wang2024yolov9}, comprises the detection phase. A compact convolutional transformer \cite{hassani2021escaping} from \texttt{fast-plate-ocr} reads the text on the license plate; we specifically used \texttt{cct-xs-v1-global-model} \cite{andresFastPlateOCR2025}.

We relied fully on LLMs to implement SPAR, specifically ChatGPT 5.1 and Claude 4 Sonnet. We provided high-level objectives as prompts (e.g. "Make a data labeling script that enables a human to select the locations of all license plate corners"). Our two methodological innovations were both suggested by LLMs. First, SPAR accounts for perspective transformations by converting the manually labeled corner data into a homography matrix and applying the adversarial patch as a rim around the license plate during training. Second, total variation loss \cite{rudin1992nonlinear} helps regularize the patch, reducing overfitting despite the small size of our training dataset. Both these techniques are well established in prior computer vision literature, though our work is the first time they have been incorporated into an adversarial attack on an ALPR system.

\subsection{Data Collection and Preprocessing}

We collected a small training dataset to model our low-resource constraint. We drove the test vehicle to several urban and suburban locations during daytime, evening, and nighttime hours, and captured images of the vehicle's license plates in an unstructured fashion, attempting to construct an intuitively diverse dataset. We developed an assisted manual labeling software to label the corners $C_{plate}$ of the license plate in each image, observing that each image could be accurately labeled in under 3 seconds. The full dataset of 300 images was labeled by a single person within several hours. 20\% of this data was held out in a validation set, so the final training dataset consisted of only 240 images.

\subsection{Patch Application}

Suppose we have input image $I \in \mathbb{R}^{C \times H \times W}$, where $C \times H \times W$ are the dimensions. Suppose we also have adversarial patch $P \in \mathbb{R}^{C \times h \times w}$, where $C \times h \times w$ are the patch dimensions. From the manually labeled license plate corners $C_{plate}$, we compute $C_{rim}$ by scaling the quadrilateral up from its geometric center by a factor of $1.6$. We also define $C_{patch}$ as the canonical corners of the patch image. We then define the homography matrix $M_{rim} : C_{patch} \rightarrow C_{rim}$ and iteratively run \texttt{inverse\_warp} over all pixels corresponding to the input image to generate a transformed adversarial patch on the same perspective as the license plate (the standard \texttt{warp\_perspective} algorithm).

\texttt{warp\_perspective($P, M_{rim}, H, W$)} returns $P'$ of the same size as the input image, where $P'_{c, y, x}$ is $0$ if $(x, y)$ is outside the rectangular bounds defined by $C_{patch}$, or equal to the roughly corresponding $P_{c, v, u}$ otherwise. We then merge $P'$ with the original image such that the rim is overlaid on the original image, but the original license plate is still visible ("cut out" of the patch). We define a white patch $P_W \in \mathbb{R}^{C \times h \times w}$ and $M_{plate} : C_{patch} \rightarrow C_{plate}$ similarly to $M_{rim}$. We then use \texttt{warp\_perspective} to create a mask defining which pixels in the image $I$ are part of the rim, which are part of the license plate, and which are part of neither:

\begin{equation}
    \mathcal{M}_{rim} = \texttt{warp\_perspective}(P_W, M_{rim}, H, W)
\end{equation}

\begin{equation}
    \mathcal{M}_{plate} = \texttt{warp\_perspective}(P_W, M_{plate}, H, W)
\end{equation}

\begin{equation}
    \mathcal{M}_{final} = \texttt{clamp}(\mathcal{M}_{rim} - \mathcal{M}_{plate}, 0, 1)
\end{equation}

With this $\mathcal{M}_{final}$, we can composite an output $I'$ with the rim applied around the license plate in the image:

\begin{equation}
    I' = I \odot (1 - \mathcal{M}_{final}) + \rho \cdot P' \odot \mathcal{M}_{final}
\end{equation}

Where $\odot$ is the Hadamard product and $\rho \in \mathbb{R}$ is a random darkening factor between 0 and 0.2 used to encourage robustness to lighting variations. We used patch dimensions $h = 256, w = 512$, and image dimensions were $H = 384, W = 384$ with both images using $C = 3$.

\subsubsection{Loss Function}

Let $\mathcal{D} : \mathbb{R}^{C \times H \times W} \rightarrow (\mathbb{R}^{C \times H \times W}, \mathbb{R})$ be the detection model and let $\mathcal{O} : \mathbb{R}^{C \times H_O \times W_O} \rightarrow \mathbb{R}^{L \times V}$ be the OCR model, where $H_O$ and $W_O$ are the input dimensions of the OCR model, $L$ is the string buffer length of the OCR output, and $V$ is the vocabulary size of the OCR output. Suppose $(B, C_{onf}) = \mathcal{D}(I')$ are detection region mask and confidence outputted by $\mathcal{D}$ on the patched image, and $A$ is the smallest rectangular bounding area encompassing all of $C_{rim}$. The first component of the loss function is the negative Intersection-over-Union between $A$ and $B$ \cite{yuUnitBoxAdvancedObject2016}, divided by the confidence value. This aims to expand the bounding box output of YOLO-v9 to cover the license plate rim and improve the rim's ability to impact the downstream OCR step. Dividing by the confidence value also prioritizes reducing detection confidence. With $\delta = 10^{-8}$ as a numerical stability constant, detection loss $\mathcal{L}_{det}$ is defined as:

\begin{equation}
    \mathcal{L}_{det} = -\frac{B \cap A}{B \cup A} \cdot \frac{1}{C_{onf} + \delta}
\end{equation}

We resize $B$ into $B' \in \mathbb{R}^{C \times H_O \times W_O}$ to fit the OCR model's input size; $O = \mathcal{O}(B')$ are the output character probabilities of the OCR model. Define $T \in \{0, 1\}^{L \times V}$ as the one-hot encoded impersonation target. $T_{smooth} = T ( 1 - \epsilon) + \frac{\epsilon}{V}$ is a smoothed label tensor with smoothing factor $\epsilon = 0.01$ that helps prevent exploding logits and gradients (this had been added by an LLM without disclosure to the operator). The probability of the true class $p_t$ is:

\begin{equation}
    p_t = \sum_{v=1}^VT_{smooth, v} \cdot O_v
\end{equation}

From here, we can calculate the focal categorical cross-entropy loss \cite{linFocalLossDense2018}, where $\alpha = 0.25$ is the balancing factor and $\gamma = 2.0$ is the focusing parameter:

\begin{equation}
    \mathcal{L}_{focal} = \sum_{l = 1}^L -\alpha (1 - p_t^{(l)})^\gamma \log(p_t^{(l)} + \delta)
\end{equation}

We rescale this by first calculating $\mathcal{L}_{baseline}$ representing the focal categorical cross-entropy loss on the dataset without any patch applied. The total OCR loss for patch $P$ is:

\begin{equation}
    \mathcal{L}_{OCR} = \frac{\mathcal{L}_{focal}}{\mathcal{L}_{baseline}}
\end{equation}

The final loss component and our work's second core contribution is a total variation (TV) regularization term \cite{rudin1992nonlinear}. This is applied only to the adversarial patch $P \in \mathbb{R}^{C \times h \times w}$, intending to encourage neighboring pixels to take similar values. This both improves print accuracy by minimizing large color changes between pixels, and also encourages the discovery of large, complex structures. We normalize by dividing by the number of comparisons.

\begin{equation}
    \begin{split}
        TV_h = \sum_{c=1}^C\sum_{i=1}^h\sum_{j=1}^{w-1} (P_{c,i,j+1} - P_{c,i,j})^2 \\
        TV_v = \sum_{c=1}^C\sum_{i=1}^{h-1}\sum_{j=1}^w (P_{c,i+1,j} - P_{c,i,j})^2 \\
        \mathcal{L}_{TV} = \frac{TV_h + TV_v}{C \cdot (h(w-1) + (h-1)w)}
    \end{split}
\end{equation}

The compound cost function $\mathcal{L}$ is thus defined as follows:

\begin{equation}
    \mathcal{L} = \frac{\mathcal{L}_{det} + \mathcal{L}_{OCR}}{2} + \zeta \cdot \mathcal{L}_{TV}
\end{equation}

Here, $\zeta = 2.5$ is a scaling factor tuned to encourage regularization without overwhelming the remaining loss components.

\subsubsection{Gradient Descent}

Adversarial patch $P$ is initialized via the Xavier method \cite{glorotUnderstandingDifficultyTraining2010}. We optimize $P$ by minimizing the cost function $\mathcal{L}$ via gradient descent. All perspective warping operations are differentiably computed by the \texttt{kornia} software package version 0.8.1 \cite{ribaKorniaOpenSource2019} to ensure partial derivatives can be computed for each patch parameter. We use the AdamW optimizer \cite{loshchilovDecoupledWeightDecay2019} to calculate the value of $P_{c, y, x}^{(t + 1)}$. We found that learning rate $\eta = 0.1$ worked best for our use case, and we used a learning rate schedule that would halve $\eta$ if no improvement in validation-set loss was detected for 5 epochs. We also used an early stopping system that would halt training when no validation-set loss improvement was detected for 20 epochs; when both homography-aware training and TV loss were enabled, this condition was never triggered.

We ran the optimization algorithm for 100 epochs, and in each epoch processed all 240 images in the training dataset, grouped into batches of size 64. We optimized twice, once with a random impersonation target to maximize disruption, and a second time with an impersonation target differing from the true license plate number by only two characters, producing $P_{disrupt}$ and $P_{impersonate}$, respectively.

\subsection{Evaluation}

We use attack success rate (ASR) to quantify the success of $P_{impersonate}$. For $P_{disrupt}$, due to the high proportion of failed detections in both physical- and digital-world evaluations even with no patch applied, we report the percent reduction in correct read rate rather than the proportion of images with a misread plate. $C_D \in [0, 1]$ is the confidence of the detection model after patch application, $ED_T \in \{0, \dots, 7\}$ is the edit distance between \texttt{fast-alpr}'s output and the true license plate number, and $ED_I \in \{0, \dots, 7\}$ is the edit distance between \texttt{fast-alpr}'s output and the attacker-defined impersonation target.

\subsubsection{Physical World:}

We contacted a professional print studio to print both $P_{disrupt}$ and $P_{impersonate}$ on a heavyweight matte poster of $8.4\text{"} \times 16.8\text{"}$ dimensions. This was the most expensive part of the attack, costing roughly \$50. We visually aligned this poster with our license plate and mounted it between the plate and vehicle, ensuring that no part of the license plate area was obscured.

We first conducted a vertical viewing angle test in full sunlight for $P_{disrupt}$, $P_{impersonate}$, and no patch (control). We used a DJI Mini camera drone in an empty parking lot and recorded distances and altitudes using its built-in GPS and altimeter. We sampled in a grid pattern distances ranging from 1 to 10 meters and altitudes ranging from 0.5 to 3.5 meters, in intervals of 1 and 0.5 meters along distance and altitude axes. We then conducted a set of horizontal viewing angle tests with a standard iPhone 15 camera from the author's height of approximately $6\text{'}$. A grid pattern of $45\text{'} \times 40\text{'}$ was established for data collection. Images were captured at $5\text{'}$ intervals in the longitudinal direction, beginning $5\text{'}$ from the license plate and extending to $50\text{'}$. At each distance, additional images were taken at $5\text{'}$ lateral intervals, ranging from $20\text{'}$ to the right of the plate to $20\text{'}$ to the left, including a centered position. We took images in multiple lighting conditions in full sunlight, dusk, night with dim street lighting, and night with camera flash (Fig. 1). The night with flash case evaluated the patches' performance when the license plate was retroreflectively illuminated, and presents a particularly difficult case in which we expected failure.

\begin{figure}
    \centering
    \includegraphics[width=1\linewidth]{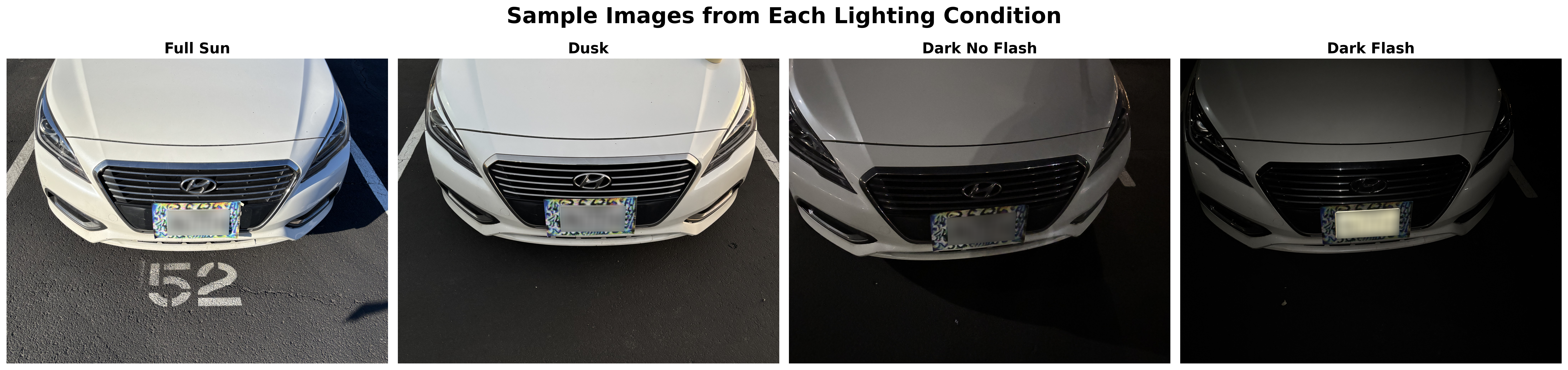}
    \caption{Sample images from each lighting condition.}
    \label{fig:placeholder}
\end{figure}

\subsubsection{Digital World:}

We evaluated further in the digital world to benchmark best-case performance and understand the specific roles of our contributions via an ablation test. We used the control images of our horizontal viewing angle test, with no patch applied, as our test set, totaling 358 images at known grid coordinates. We manually labeled the corners of the license plates in these images, then digitally inserted $P_{disrupt}$ and $P_{impersonate}$ and ran \texttt{fast-alpr}. We finally conducted an ablation test dropping out homography-aware training, TV loss, and both, and compared each variant against the full SPAR approach.

\subsubsection{Distance Correlation Analysis:}

To quantify whether SPAR’s performance depends on camera geometry, we measured statistical dependence between performance metrics and camera distance/angle using distance correlation. Distance correlation ranges from $0$ to $1$ and detects arbitrary nonlinear dependence, equaling zero if and only if the variables are independent \cite{szekelyMeasuringTestingDependence2007}. For performance metric $X$ and camera position $Y$, we compute distance correlation $\mathrm{dCor}(X, Y)$.

To determine plausible natural variation of $\mathrm{dCor}$ in independent data, we construct an empirical null distribution $\mathcal{D}_{null}$ by generating 10,000 random permutations of the camera position parameter $Y$ across samples, preserving marginal distributions while destroying dependence. For noise threshold $\alpha$ of $\mathcal{D}_{null}$, we define the independence noise floor $\delta_\alpha$ as the $\alpha$-quantile of $\mathcal{D}_{null}$; we use this strictly to contextualize dCor, not as a statistical test. To express effect magnitude relative to the independence noise floor, we report the normalized dependence ratio:

\begin{equation}
R_\alpha = \frac{\mathrm{dCor}(X,Y)}{\delta_\alpha}.
\end{equation}

To assess attack-induced dependence, we compare raw distance correlations and normalized dependence ratios between the attack type and control with $\Delta \text{dCor}$ and $\Delta R_\alpha$. Images with no license plate detected were assigned confidence 0 and edit distances 7 (maximum possible). We selected the conservative noise threshold $\alpha = 0.99$ as a bound of non-meaningful levels of statistical dependence. We represent camera distance as $d$ and camera angle as $\theta$.

\section{Results}

\begin{figure}
    \centering
    \includegraphics[width=1\linewidth]{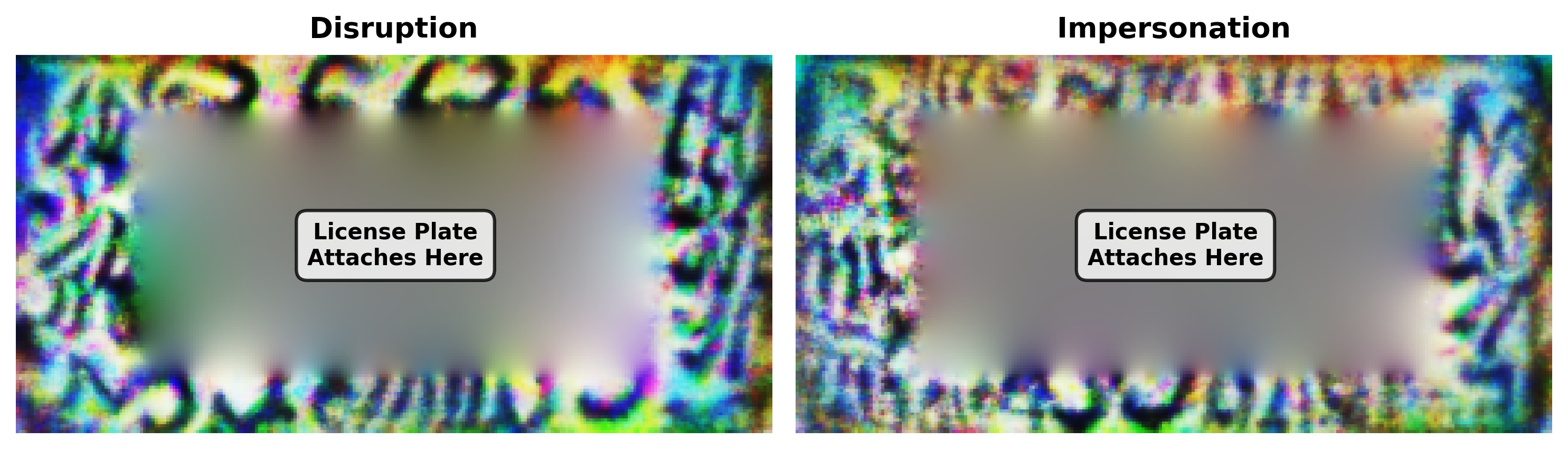}
    \caption{Final adversarial rims for disruption and impersonation objectives.}
    \label{fig:placeholder}
\end{figure}

The final trained patches (Fig. 2) exhibited fascinating large-scale structure, including characters resembling "w", "3", "M", and East Asian script.

\subsection{Physical World}

Across all physical world images, we categorized the result of \texttt{fast-alpr} as either a correct read, incorrect read, successful impersonation, or failed detection. We found that $P_{impersonate}$ caused a 39.8\% reduction in correct reads and $P_{disrupt}$ caused a 49.2\% reduction; $P_{impersonate}$ achieved an impersonation ASR of 13.8\%. Additionally, $P_{disrupt}$ achieved a mean confidence reduction of 17.8\%, and $P_{impersonate}$ achieved a mean confidence reduction of 3.2\%.

For the horizontal viewing angle tests, we broke down patch performance by lighting condition (Fig. 3). Due to the large distances at which many images were taken during the horizontal viewing angle test, the \texttt{yolo-v9-t-384-license\\-plate-end2end} detection model often failed to detect the license plate even in the control case. We thus opted to use the larger \texttt{yolo-v9-t-608-license-plate-end2end} detection model to ensure proper evaluation of the OCR portion of the attack. This choice introduced a strictly more challenging evaluation setting.

\begin{figure}
    \centering
    \includegraphics[width=1\linewidth]{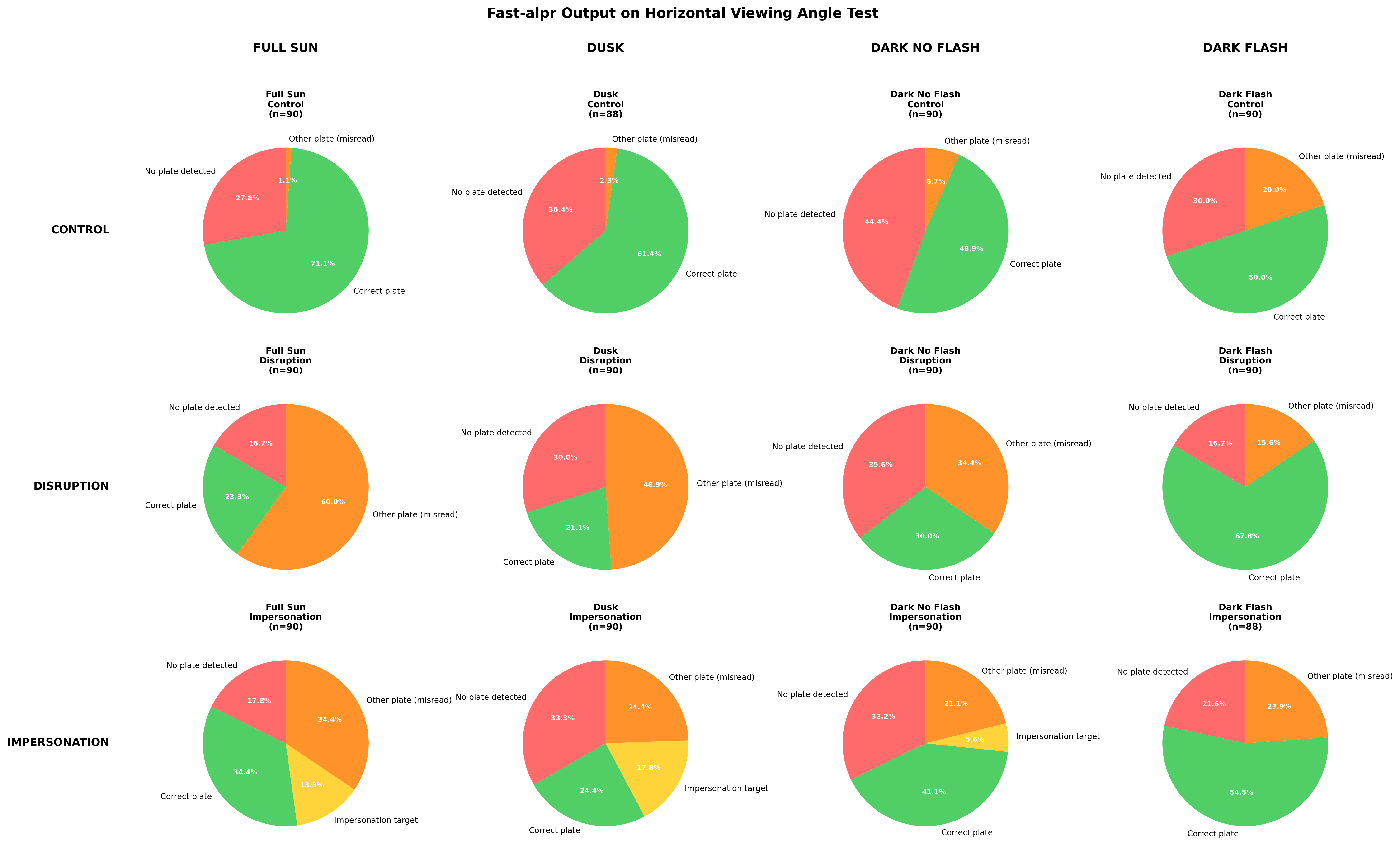}
    \caption{\texttt{fast-alpr} output on horizontal viewing angle test.}
    \label{fig:placeholder}
\end{figure}

$P_{disrupt}$ achieves optimal performance in full sunlight, inducing a 60.0\% reduction in correct reads. $P_{impersonate}$ achieves optimal performance in dusk lighting, inducing \texttt{fast-alpr} to output the impersonation target 17.8\% of the time. Both attacks greatly struggled on the dark flash case because of the license plate's retroreflectivity. Although we could have corrected for this effect by printing our border onto retroreflective metal sheeting, we chose not to do so to better understand the failure mode. Interestingly, $P_{disrupt}$ appears to have \textit{improved} the performance of \texttt{fast-alpr} in the dark flash condition, raising the rate of correct reads by 35\%, reducing the number of failed detections by 44.3\%, reducing the number of misreads by 22.0\% (Fig. 3), and improving the detection confidence above the control case (Fig. 4). We explore this further in our Discussion.

\begin{figure}[t]
    \centering
    \includegraphics[width=\linewidth]{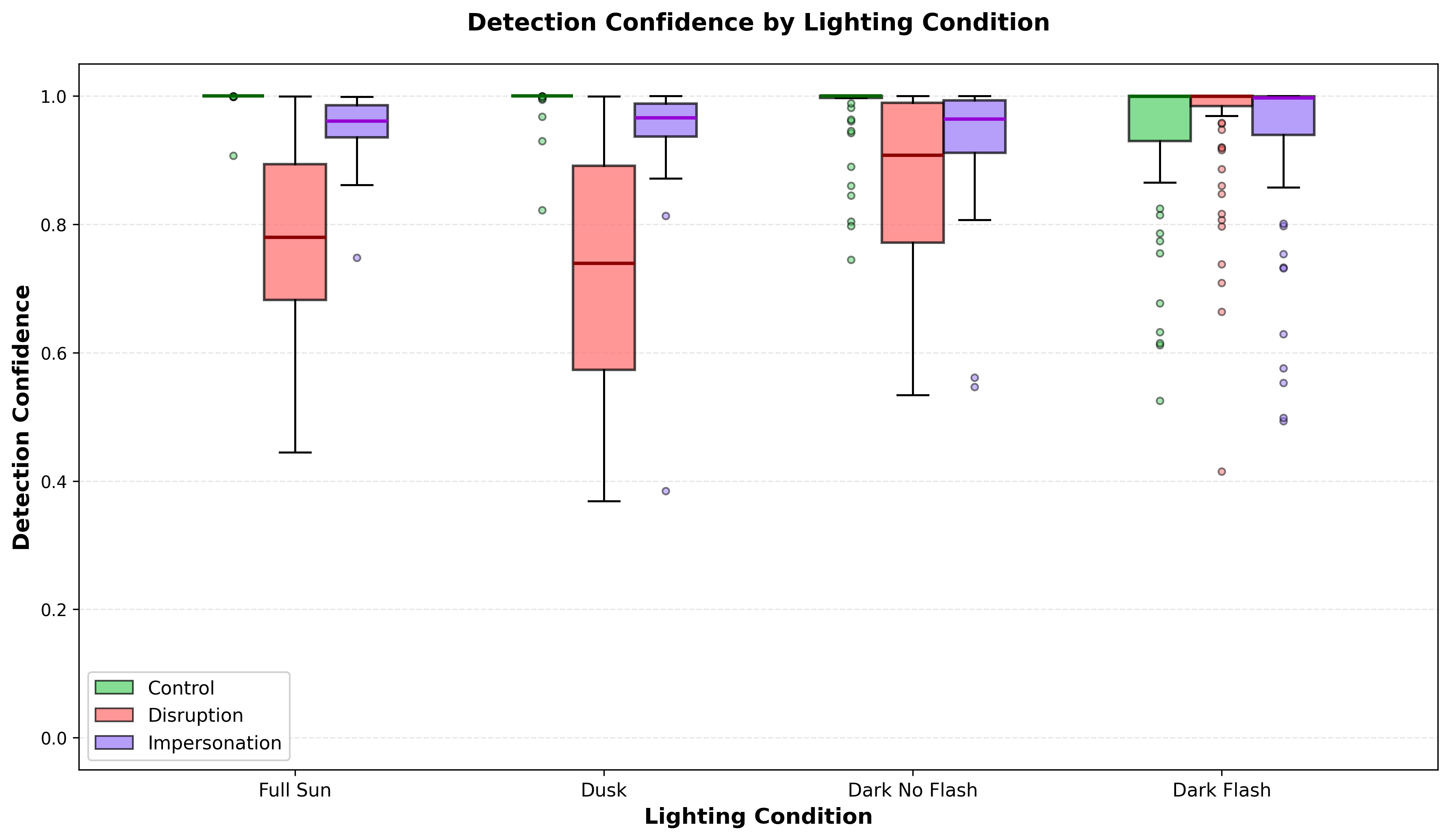}
    \caption{Detection model's confidence distributions under various lighting conditions.}
    \label{fig:confidence_lighting}
\end{figure}

\begin{figure}[t]
    \centering
    \includegraphics[width=\linewidth]{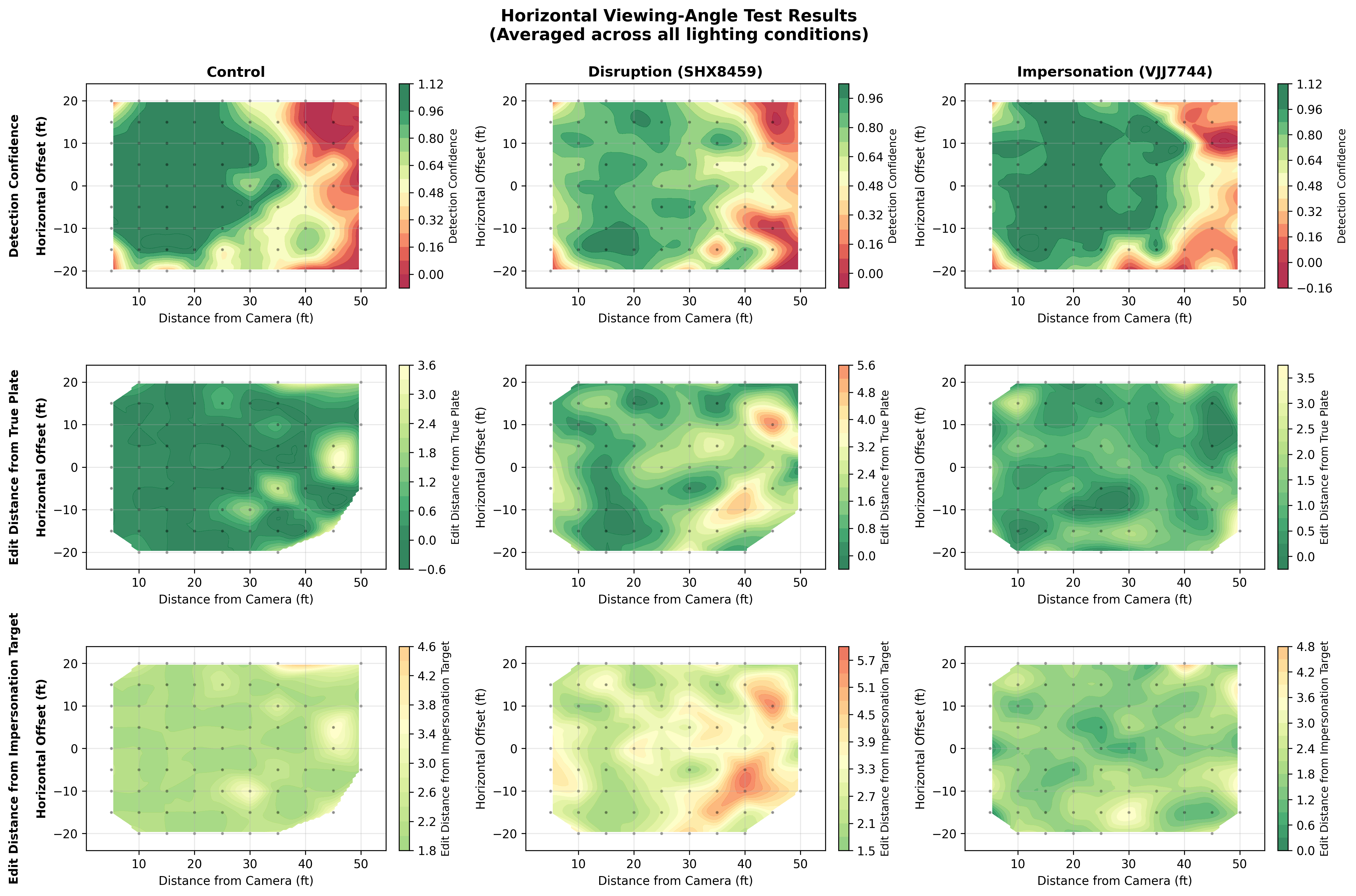}
    \caption{Horizontal viewing-angle test results (metrics averaged over all lighting conditions).}
    \label{fig:horizontal_angle}
\end{figure}

For the horizontal viewing angle tests, we created contour plots showing how the detection confidence and edit distances changed over the grid pattern (Fig. 5). Images without a detected license plate were assigned 0 confidence and excluded from edit distance contour calculations. No obvious correlation is apparent between SPAR's performance and viewing angle, although there does appear to be a dependence on distance stemming from the model itself. To quantify this, we conducted our distance correlation analysis (also including vertical viewing angle results).

\begin{table}[t]
\centering
\caption{Dependence between performance metrics and camera parameters with no patch applied. $R_\alpha$ denotes distance correlation normalized by the permutation-derived independence bound. Vertical viewing-angle test data included.}
\label{tab:control_dependence}
\begin{tabular}{llccc}
\toprule
$X$ & $Y$ & dCor & $\delta_\alpha$ & $R_\alpha$ \\
\midrule
$C_D$  & $d$      & 0.5342 & 0.1314 & 4.07 \\
$C_D$  & $\theta$ & 0.1350 & 0.1286 & 1.05 \\
$ED_T$ & $d$      & 0.5434 & 0.1297 & 4.19 \\
$ED_T$ & $\theta$ & 0.1378 & 0.1300 & 1.06 \\
$ED_I$ & $d$      & 0.5314 & 0.1312 & 4.05 \\
$ED_I$ & $\theta$ & 0.1321 & 0.1278 & 1.03 \\
\bottomrule
\end{tabular}
\end{table}

\begin{table*}[t]
\centering
\caption{Attack-induced changes in dependence relative to no patch applied.
$\Delta \mathrm{dCor}$ and $\Delta R_\alpha$ are computed as (attack $-$ control)
for each metric--camera parameter pair.}
\label{tab:attack_deltas}

\begin{minipage}{0.48\textwidth}
\centering
\textbf{Disruption Attack}

\begin{tabular}{lllrrr}
\toprule
$X$ & $Y$ & $\mathrm{dCor}$ & $R_\alpha$ & $\Delta \mathrm{dCor}$ & $\Delta R_\alpha$ \\
\midrule
$C_D$  & $d$      & 0.5185 & 3.58 & $-0.1084$ & $-0.82$ \\
$C_D$  & $\theta$ & 0.1206 & 0.84 & $-0.0053$ & $-0.08$ \\
$ED_T$ & $d$      & 0.4754 & 3.33 & $-0.1525$ & $-1.24$ \\
$ED_T$ & $\theta$ & 0.1109 & 0.76 & $+0.0147$ & $+0.08$ \\
$ED_I$ & $d$      & 0.4567 & 3.13 & $-0.1678$ & $-1.30$ \\
$ED_I$ & $\theta$ & 0.1172 & 0.82 & $+0.0125$ & $+0.05$ \\
\bottomrule
\end{tabular}
\end{minipage}
\hfill
\begin{minipage}{0.48\textwidth}
\centering
\textbf{Impersonation Attack}

\begin{tabular}{lllrrr}
\toprule
$X$ & $Y$ & $\mathrm{dCor}$ & $R_\alpha$ & $\Delta \mathrm{dCor}$ & $\Delta R_\alpha$ \\
\midrule
$C_D$  & $d$      & 0.5659 & 3.99 & $-0.0100$ & $-0.02$ \\
$C_D$  & $\theta$ & 0.1540 & 1.07 & $-0.0287$ & $-0.24$ \\
$ED_T$ & $d$      & 0.5400 & 3.74 & $-0.0573$ & $-0.42$ \\
$ED_T$ & $\theta$ & 0.1589 & 1.10 & $-0.0097$ & $-0.08$ \\
$ED_I$ & $d$      & 0.5963 & 4.11 & $-0.0073$ & $-0.15$ \\
$ED_I$ & $\theta$ & 0.1569 & 1.10 & $-0.0172$ & $-0.15$ \\
\bottomrule
\end{tabular}
\end{minipage}

\end{table*}

These results indicate that the \texttt{fast-alpr} base model's performance already has a high level of dependence on $d$ and a weak-to-insignificant dependence on $\theta$ (Table 2). $P_{impersonate}$ moderately reduced this dependence, with the notable reductions observed in $\mathrm{dCor}(C_D, \theta)$, $\mathrm{dCor}(ED_T, \theta)$, $\mathrm{dCor}(ED_I, d)$, and $\mathrm{dCor}(ED_I, \theta)$. $P_{disrupt}$ caused much stronger reductions in $\mathrm{dCor}(C_D, d)$ and $\mathrm{dCor}(ED_T, d)$, but marginally increased $\mathrm{dCor}(ED_T, \theta)$. Additionally, $P_{impersonate}$'s correlations with viewing angle are all below the significance threshold $\delta_\alpha$, and $P_{disrupt}$'s correlations with the same are either below or slightly above, so we conclude that SPAR has succeeded in its attack with negligible dependence on camera geometry (Table 3).

To further study the generalizability of our approach in the physical world, we compared the detection confidence of \texttt{yolo-v9-t-384-license-plate-end2end} (small detector) vs \texttt{yolo-v9-t-608-license-plate-end2end} (large detector) when presented with each adversarial patch. We found that $P_{disrupt}$ reduces the small detector's confidence by 20.5\% and the large detector's confidence by 14.2\%, on average. In ideal (full sun) lighting conditions, those reductions increase to 31.2\% and 22.0\%, respectively. Thus SPAR successfully generalizes to larger variants of YOLOv9.

\subsection{Digital World}

We found that with digital application of the patch, \texttt{fast-alpr}'s correct read frequency decreased by 72.5\% with $P_{disrupt}$ and 56.7\% with $P_{impersonate}$ applied. The impersonation attack was successful in 15.9\% of cases, and the disruption attack induced misreads in 46.6\% of cases (excluding cases where no plate was detected). $P_{disrupt}$ reduced the detection model's confidence by 23.4\% to 0.766 and $P_{impersonate}$ reduced the confidence by 3.8\% to 0.962, on average.

We conducted an ablation test of the primary contributions of our work: homography-aware patch insertion during training and the total variation loss component for regularization and structural pattern discovery. We retrained SPAR with each component removed individually, as well as with both removed, and evaluated all variants using the same digital-world evaluation framework to isolate their respective effects. From the final patches of each case (Fig. 6), it is visibly apparent that TV loss achieves its objective of regularizing the pixel values into more natural patterns. Homography transformations, too, promote the discovery of larger patterns rather than exploiting a guaranteed pixel positioning in the OCR input when patches are applied rectangularly.

\begin{figure}
    \centering
    \includegraphics[width=0.75\linewidth]{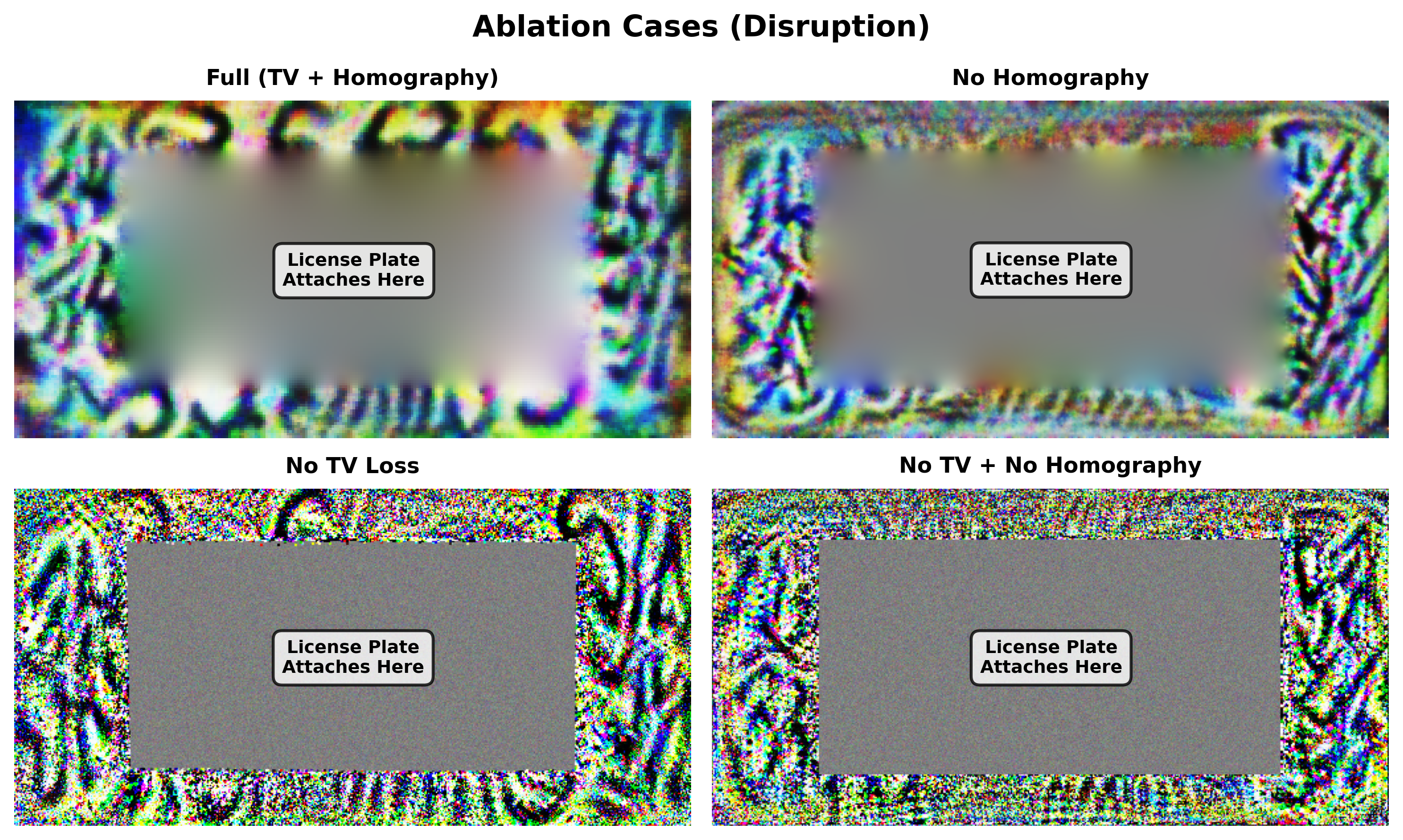}
    \caption{All final patches used in ablation study.}
    \label{fig:placeholder}
\end{figure}

We found that both TV loss and homography-aware training are critical for SPAR's performance. Removing TV loss causes a 98.7\% increase in correct reads on the disruption attack and near complete failure at the impersonation attack. Removing homography-based patch application during training causes a 107\% increase in correct reads and a 70.4\% reduction in successful impersonations. Removing both results in near-identical performance to removing TV loss alone on the impersonation attack, and slightly better performance than removing TV loss alone on the disruption attack (Fig. 7). We believe this is due to the regularization effect of TV loss, without which SPAR tends to overfit to its training data as a simpler solution to the complex problem presented by homography-based insertion. We observed much greater divergence between training and validation loss when TV loss was excluded, and worse final validation-set performance than when homography-aware training was also excluded (Fig. 8).

\begin{figure}[t]
\centering

\begin{minipage}{0.48\linewidth}
    \centering
    \includegraphics[width=\linewidth]{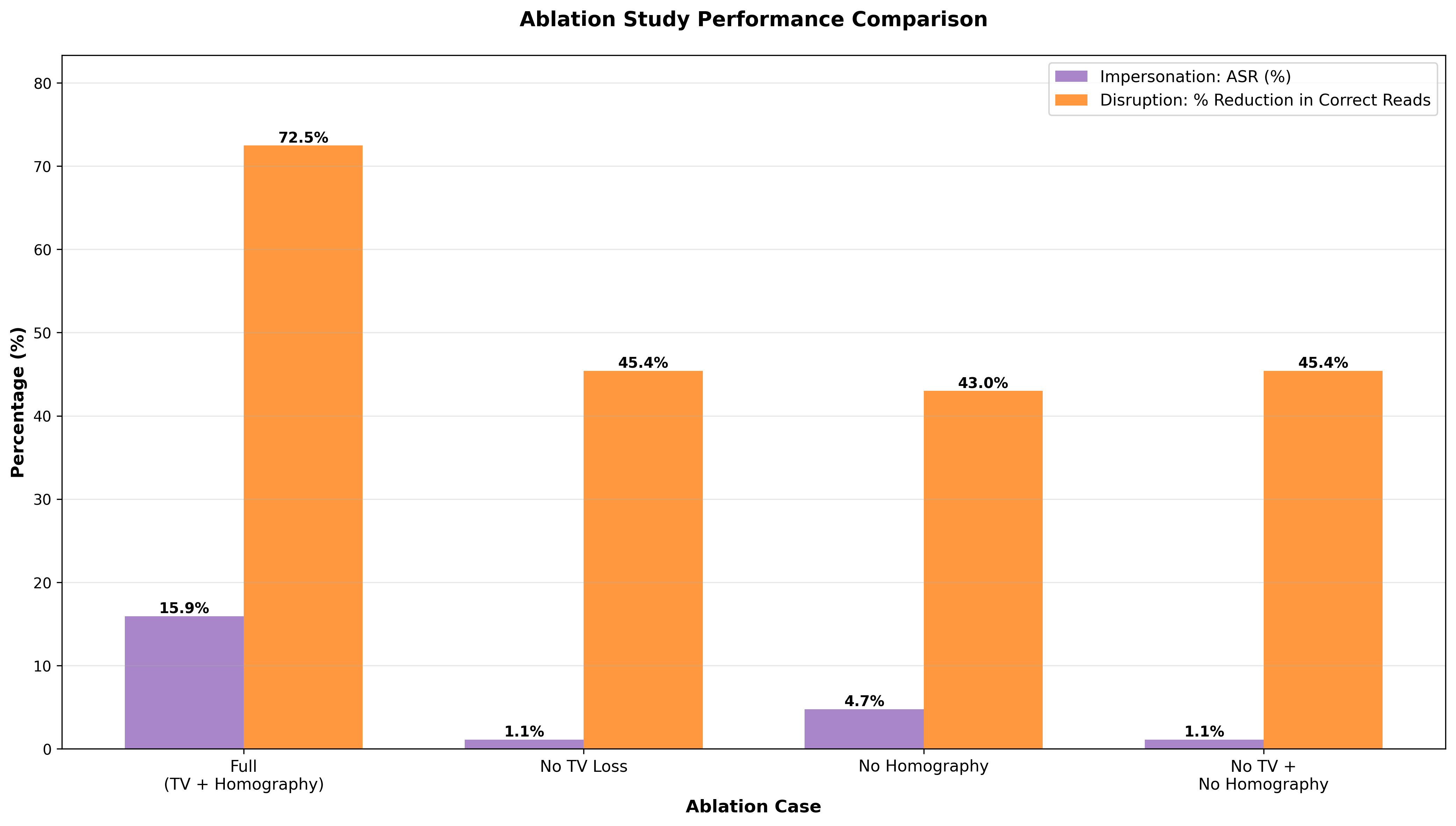}
    \caption{Performance of each ablation case compared to standard SPAR.}
    \label{fig:ablation_pie}
\end{minipage}
\hfill
\begin{minipage}{0.48\linewidth}
    \centering
    \includegraphics[width=\linewidth]{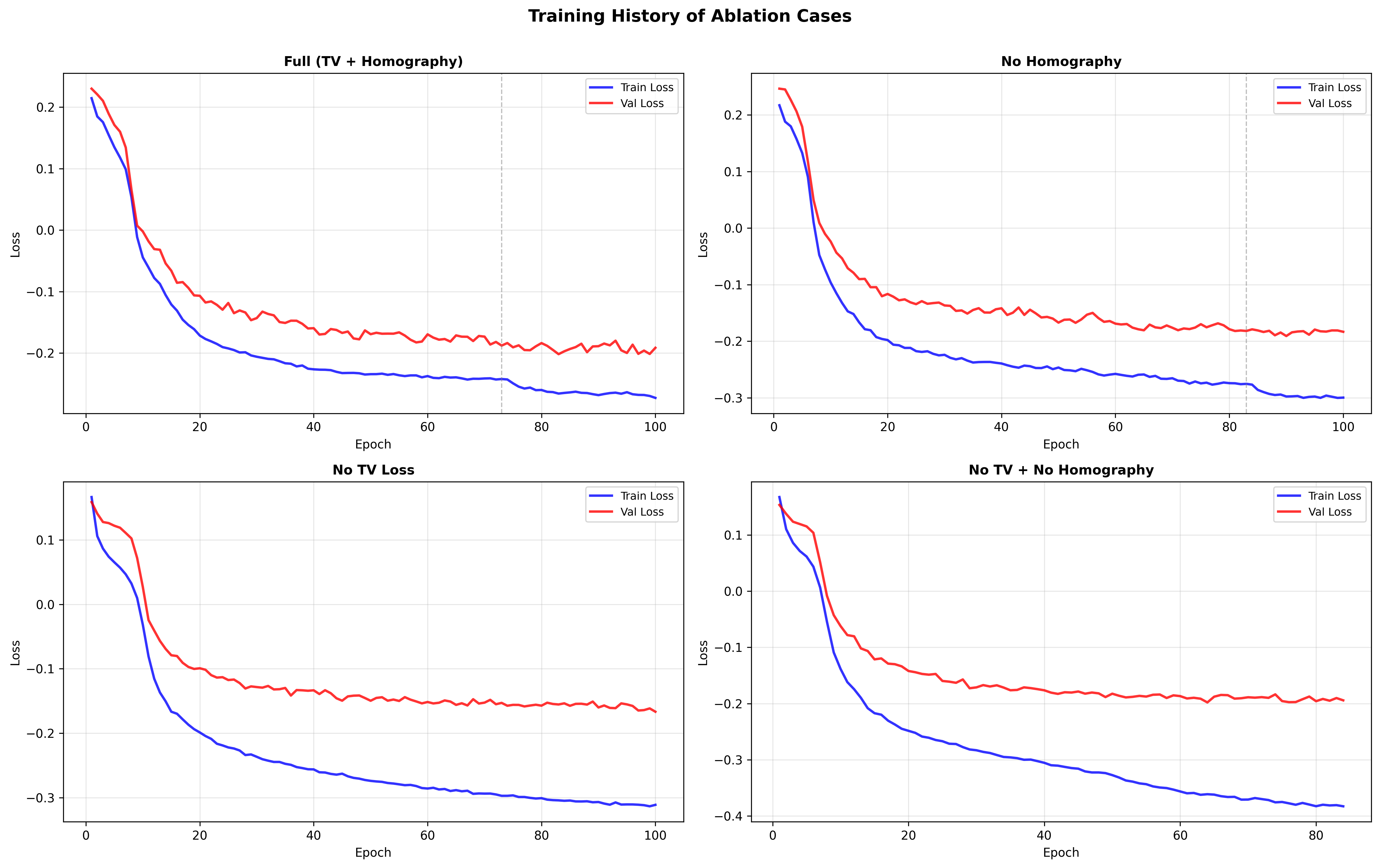}
    \caption{Training and validation loss curves during ablation testing for the disruption objective.}
    \label{fig:ablation_training}
\end{minipage}

\end{figure}

Finally, to clarify the specific roles of homography-aware training and TV loss in promoting resistance to viewing angle and camera distance transformations, we repeated the distance correlation test procedure, calculating $\Delta R_\alpha$ in reference to $R_\alpha$ of the full disruption and impersonation attacks. We digitally applied patches produced by the full SPAR process and calculated $R_\alpha$ values. These varied somewhat from the physical-world evaluation, likely because of differences in the dataset used, patch application method, and the removal of lighting as a factor, but we proceeded with them for consistency with the ablation test cases.

We then calculated $\Delta R_\alpha$ for all ablation cases (Table 4). For $P_{disrupt}$, we found substantial increases in $\mathrm{dCor}$ on both distance and viewing angle when homography-aware training was removed. Weaker, but still positive effects on $\mathrm{dCor}$ were observed for dropping out TV loss alone, as well as TV + homography. TV loss alone seems to achieve a higher baseline level of attack efficacy, and homography-aware training may squeeze out extra performance by addressing off-angle, longer-distance cases. For $P_{impersonate}$, TV loss is required for impersonation attack efficacy (Fig. 7), but appears to be more sensitive to optimal camera parameters given the increased $\mathrm{dCor}$ when homography-aware training is dropped out. Similar to the disruption attack, it is likely that SPAR's success where the naive baseline fails stems from the ability to synthesize large-scale patterns and make those patterns robust to camera distance and viewing angle transformations.

\begin{table*}[t]
\centering
\caption{Ablation-induced changes in normalized dependence relative to the full attack.
$\Delta R_\alpha$ is computed as (ablation $-$ original SPAR) for each metric--parameter pair.}
\label{tab:ablation_deltas}

\begin{minipage}{0.48\textwidth}
\centering
\textbf{Disruption Attack}

\begin{tabular}{lllrr}
\toprule
Removed & $X$ & $Y$ & $R_\alpha$ & $\Delta R_\alpha$ \\
\midrule

Homography & $C_D$  & $d$      & 3.90 & $\DeltaR{+0.32}$ \\
Homography & $C_D$  & $\theta$ & 1.17 & $\DeltaR{+0.33}$ \\
Homography & $ED_T$ & $d$      & 4.14 & $\DeltaR{+0.81}$ \\
Homography & $ED_T$ & $\theta$ & 1.17 & $\DeltaR{+0.41}$ \\
Homography & $ED_I$ & $d$      & 3.94 & $\DeltaR{+0.81}$ \\
Homography & $ED_I$ & $\theta$ & 1.20 & $\DeltaR{+0.38}$ \\

TV Loss & $C_D$  & $d$      & 3.84 & $\DeltaR{+0.26}$ \\
TV Loss & $C_D$  & $\theta$ & 0.80 & $\DeltaR{-0.04}$ \\
TV Loss & $ED_T$ & $d$      & 3.87 & $\DeltaR{+0.54}$ \\
TV Loss & $ED_T$ & $\theta$ & 0.81 & $\DeltaR{+0.05}$ \\
TV Loss & $ED_I$ & $d$      & 3.71 & $\DeltaR{+0.58}$ \\
TV Loss & $ED_I$ & $\theta$ & 0.79 & $\DeltaR{-0.03}$ \\

Both & $C_D$  & $d$      & 3.74 & $\DeltaR{+0.16}$ \\
Both & $C_D$  & $\theta$ & 0.93 & $\DeltaR{+0.09}$ \\
Both & $ED_T$ & $d$      & 3.77 & $\DeltaR{+0.44}$ \\
Both & $ED_T$ & $\theta$ & 0.88 & $\DeltaR{+0.12}$ \\
Both & $ED_I$ & $d$      & 3.69 & $\DeltaR{+0.56}$ \\
Both & $ED_I$ & $\theta$ & 0.87 & $\DeltaR{+0.05}$ \\

\bottomrule
\end{tabular}
\end{minipage}
\hfill
\begin{minipage}{0.48\textwidth}
\centering
\textbf{Impersonation Attack}

\begin{tabular}{lllrr}
\toprule
Removed & $X$ & $Y$ & $R_\alpha$ & $\Delta R_\alpha$ \\
\midrule

Homography & $C_D$  & $d$      & 4.04 & $\DeltaR{+0.05}$ \\
Homography & $C_D$  & $\theta$ & 1.17 & $\DeltaR{+0.10}$ \\
Homography & $ED_T$ & $d$      & 3.92 & $\DeltaR{+0.18}$ \\
Homography & $ED_T$ & $\theta$ & 1.28 & $\DeltaR{+0.18}$ \\
Homography & $ED_I$ & $d$      & 3.98 & $\DeltaR{-0.13}$ \\
Homography & $ED_I$ & $\theta$ & 1.22 & $\DeltaR{+0.12}$ \\

TV Loss & $C_D$  & $d$      & 4.05 & $\DeltaR{+0.06}$ \\
TV Loss & $C_D$  & $\theta$ & 0.93 & $\DeltaR{-0.14}$ \\
TV Loss & $ED_T$ & $d$      & 4.12 & $\DeltaR{+0.38}$ \\
TV Loss & $ED_T$ & $\theta$ & 0.91 & $\DeltaR{-0.19}$ \\
TV Loss & $ED_I$ & $d$      & 4.07 & $\DeltaR{-0.04}$ \\
TV Loss & $ED_I$ & $\theta$ & 0.92 & $\DeltaR{-0.18}$ \\

Both & $C_D$  & $d$      & 4.14 & $\DeltaR{+0.15}$ \\
Both & $C_D$  & $\theta$ & 1.00 & $\DeltaR{-0.07}$ \\
Both & $ED_T$ & $d$      & 4.16 & $\DeltaR{+0.42}$ \\
Both & $ED_T$ & $\theta$ & 1.03 & $\DeltaR{-0.07}$ \\
Both & $ED_I$ & $d$      & 4.11 & $\DeltaR{-0.00}$ \\
Both & $ED_I$ & $\theta$ & 1.00 & $\DeltaR{-0.10}$ \\

\bottomrule
\end{tabular}
\end{minipage}

\end{table*}

\section{Discussion}

Our results broadly demonstrate SPAR's success at its core objectives. SPAR in the physical world reduced \texttt{fast-alpr}'s correct read rate by 49.2\% (60\% in full sunlight), and had an impersonation ASR of 13.8\% (17.8\% in dusk lighting). It also achieved a 72.5\% reduction in correct read rate and a 15.9\% impersonation ASR in the digital world. $P_{disrupt}$'s attacks on the detection model caused a 17.8\% confidence reduction in the physical world and 23.4\% in the digital world. SPAR achieved this without internal access to ALPR software or hardware during attack execution, without obscuring or altering the primary license plate area, and without creating excessive dependence on camera distance or viewing angle. The attack was carried out for a total cost of under \$100, with an approach designed and fully implemented by agentic coding tools. SPAR thus satisfies its constraints of realistic access, street-legality, simplicity, and physical-world practicality.

Adversarial patches function by inducing large-magnitude feature norms in their target models, overwhelming legitimate signal \cite{yuDefendingUniversalAdversarial2021}. Our ablation study shows that the solution of least resistance discovered by simple gradient descent optimization relies on high-magnitude input pixels with high contrast to their neighbors to induce these large-magnitude feature norms. However, these patches fail to generalize to the evaluation dataset when applied via homography. Conversely, our most successful patches exhibit large, well-defined, character-like structures. We believe this reflects a change in strategy from fuzzing early convolution filters with extreme inputs, to hijacking real features extracted and used by the model to make classification decisions.

Total variation loss promotes the emergence of these structures in the final patches. It makes the impersonation attack feasible and forces SPAR to find a different strategy for increasing CNN feature norms. It also regularizes the patch and reduces overfitting during training, enabling the attack to be executed with a very small training dataset. Homography-aware training works in tandem to break the dependence on viewing geometry when TV loss is used alone. It further promotes the development of character-like structures by changing pixels' distances to their neighbors depending on the viewing angle, making it difficult to coordinate high-magnitude pixel-perfect attacks.

In the dark flash condition, $P_{disrupt}$ seemed to improve the performance of \texttt{fast-alpr}. We believe this is due to the IoU component of $\mathcal{L}_{det}$, which rewards the model for producing a bounding box surrounding the applied patch. It is likely that this was the limiting factor for \texttt{fast-alpr} to output accurate results in the control case. The retroreflectivity greatly increased the contrast of the license plate against the surrounding background, so confidence and OCR accuracy could be high once the location of the license plate in the image was determined. $P_{disrupt}$ is far better at reducing detection model confidence than $P_{impersonate}$ overall (Fig. 4), so this is likely why we see this effect much more prominently in the disruption attack than impersonation. We therefore interpret this as a partial success that could be converted into a full success were the adversarial rim printed onto retroreflective sheeting.

\subsection{Comparison to Prior Work}

\begin{table}[t]
\centering
\caption{Comparison of attack success rates (ASR) reported in prior ALPR attacks. For SPAR’s $P_{\text{disrupt}}$, reduction in correct reads is reported instead of ASR.}
\label{tab:asr_comparison}
\begin{tabular}{lcccc}
\toprule
\textbf{Approach} 
& \makecell{\textbf{Disrupt} \\ \textbf{(Digital)}} 
& \makecell{\textbf{Impers.} \\ \textbf{(Digital)}} 
& \makecell{\textbf{Disrupt} \\ \textbf{(Physical)}} 
& \makecell{\textbf{Impers.} \\ \textbf{(Physical)}} \\
\midrule
Zha et al.        & 97.3\% & 89.5\% & N/A & 91.4\% \\
Gu et al.         & N/A    & N/A    & N/A & N/A    \\
Qian et al.       & N/A    & 93.3\% & N/A & N/A    \\
Kwon \& Baek      & 85.0\% & N/A    & N/A & N/A    \\
Jiamsuchon et al. & 91.0\% & 70.0\% & N/A & N/A    \\
Zhang et al.      & 95.6\% & N/A    & N/A & N/A    \\
\midrule
\textbf{SPAR}     & \textbf{72.5\%} & \textbf{15.9\%} & \textbf{60.0\%} & \textbf{17.8\%} \\
\bottomrule
\end{tabular}
\end{table}

We present thorough comparisons to existing literature in this space (Table 5). Because no prior work has attempted physical-world validation in challenging lighting conditions, we select SPAR's physical-world performance metrics from the optimal lighting conditions. SPAR's objective is not to produce a state-of-the-art attack on ALPR systems; it is to show what might be trivially possible with LLM assistance, and quantify the risk with physical-world evaluation. Under that lens, SPAR is a simple solution that works unexpectedly well, inviting us to consider adversarial attacks as a practical threat to ALPRs.

\subsection{Limitations, Defenses, Ethics}

SPAR is a highly focused attack against a single open-source ALPR system (\texttt{fast-alpr}) that requires full access to internal model weights during training. Broad generalization is not the purpose of our work, and we do not claim our patch designs would be successful on other license plates, other vehicles, or other ALPR systems. We instead aim to demonstrate that such an attack is possible, practical, and performant in the real world if an adversary gains access to an ALPR's internal model weights, even if they have limited technical knowledge.

Defenders can protect ALPR systems against many adversarial attacks using well-established countermeasures in the literature. Simply compressing and then re-sharpening an image (e.g. via an autoencoder) can eliminate adversarial perturbations, and clipping feature norms would likely be sufficient to protect against simple SPAR-like attacks \cite{yuDefendingUniversalAdversarial2021}. Attack detection is also possible, as most adversarial patches (including SPAR) appear very unnatural to the human eye. Spot attacks might be the most problematic, as despite being illegal, it may be difficult to determine whether dirt marks on a license plate are due to an adversarial attack or an offroad trip; including adversarial examples in the training data can improve robustness in these cases \cite{shresthaAdversarial2023}.

\subsubsection{Ethics Statement}

Our finding that LLMs can automate a large portion of the work needed to execute a physical-world adversarial attack against a real ALPR system is deeply concerning. It is because of this finding that we have pursued publication of this work, as the ideas presented here already exist as separate, well-established elements in the prior literature, and we have demonstrated that LLMs are already capable of synthesizing them into a full exploitation solution. We have demonstrated a white-box attack against a single open-source system to raise awareness in the vehicle security community about the realistic threat of LLM-assisted adversarial attacks and the sensitivity of neural network weights, but have declined to pursue transfer to commercial systems to avoid enabling real-world harm.

Due to the risks of LLM-assisted modifications to our software enabling that transfer, we have elected not to release the software implementing our attack. We believe we have included adequate technical details in our manuscript to enable legitimate researchers to replicate our findings. Additionally, because our dataset contains sensitive personally-identifiable information such as an author's real license plate, we are unable to release the dataset used to train and evaluate SPAR.

\section{Conclusion}

SPAR is a street-legal physical-world attack that disrupted both the detection and OCR portions of \texttt{fast-alpr} simultaneously, while being robust to camera perspective transformations. SPAR achieves this with a combination of homography-aware training and total variation loss, which promote the emergence of large-scale character-like structures that hijack features used by \texttt{fast-alpr} to read license plates. Despite being a white-box attack against a single system, our attack model could be repeated by a low-resourced adversary with LLM assistance. Our results demonstrate that such adversarial attacks are no longer a matter of theory and must be part of the security model for modern ALPR systems. We hope future work can address key limitations of SPAR, including retroreflectivity, infrared-spectrum cameras, and black-box training and evaluation, as well as develop more robust defenses against this form of attack.

\bibliographystyle{splncs04}
\bibliography{References}
\end{document}